\documentclass[10pt,twocolumn,letterpaper]{article}

\usepackage[pagenumbers]{cvpr} 

\definecolor{cvprblue}{rgb}{0.21,0.49,0.74}
\definecolor{ibmred}{rgb}{0.905,0.435,0.317}
\definecolor{mygray}{rgb}{.929, .929, .913}
\definecolor{mygreen}{RGB}{0,128,0} 
\definecolor{myred}{RGB}{255,0,0}   
\usepackage[pagebackref,breaklinks,colorlinks,allcolors=cvprblue]{hyperref}

\usepackage{amssymb}
\usepackage{amsmath}

\usepackage{hyperref}       
\usepackage{url}            
\usepackage{booktabs}       
\usepackage{amsfonts}       
\usepackage{nicefrac}       
\usepackage{microtype}      
\usepackage{xcolor}         

\usepackage{amsfonts}
\usepackage{multirow}
\usepackage{utfsym}
\usepackage{tabularx}
\usepackage{graphicx}
\usepackage{caption}
\usepackage{makecell}

\AtBeginDocument{
\setlength{\abovedisplayskip}{0.6ex}
\setlength{\belowdisplayskip}{0.6ex}
}
\allowdisplaybreaks[4]

\def\paperID{} 
\def\confName{CVPR}
\def\confYear{2026}

\title{SimpleMatch: A Simple and Strong Baseline for Semantic Correspondence}

\author{
    Hailong Jin$^1$ \quad
    Huiying Li$^1$\thanks{Corresponding author.} \\
    $^1$Jilin University \quad
}

\begin{document}
\maketitle

\begin{abstract}

Recent advances in semantic correspondence have been largely driven by the use of pre-trained large-scale models. However, a limitation of these approaches is their dependence on high-resolution input images to achieve optimal performance, which results in considerable computational overhead. In this work, we address a fundamental limitation in current methods: the irreversible fusion of adjacent keypoint features caused by deep downsampling operations. This issue is triggered when semantically distinct keypoints fall within the same downsampled receptive field (e.g., $16\times16$ patches). To address this issue, we present \textbf{SimpleMatch}, a simple yet effective framework for semantic correspondence that delivers strong performance even at low resolutions. We propose a lightweight upsample decoder that progressively recovers spatial detail by upsampling deep features to 1/4 resolution, and a multi-scale supervised loss that ensures the upsampled features retain discriminative features across different spatial scales. In addition, we introduce sparse matching and window-based localization to optimize training memory usage and reduce it by 51\%. At a resolution of $252\times252$ ($3.3\times$ smaller than current SOTA methods), SimpleMatch achieves superior performance with 84.1\% PCK@0.1 on the SPair-71k benchmark. We believe this framework provides a practical and efficient baseline for future research in semantic correspondence. Code is available at: \url{https://github.com/hailong23-jin/SimpleMatch}.

\end{abstract}

\section{Introduction}

Semantic correspondence, as a fundamental computer vision task, focuses on establishing visual correspondences between images sharing the same category. It plays an important role in many vision applications, such as image editing \cite{zheng2022bridging,peebles2022gan}, robot manipulation \cite{ju2024robo,xue2023useek}, and 3D reconstruction \cite{zhao2024michelangelo}. The recent emergence of powerful pre-trained vision models \cite{rombach2022high,oquab2023dinov2}  has dramatically advanced the field by providing rich, high-level feature representations that significantly enhance the matching performance. However, these methods typically require high-resolution input images, such as 840x840, to achieve optimal performance, which inevitably introduces significant computational overhead and severely limits their applicability.

\begin{figure*}
    \centering
    \includegraphics[width=0.8\linewidth]{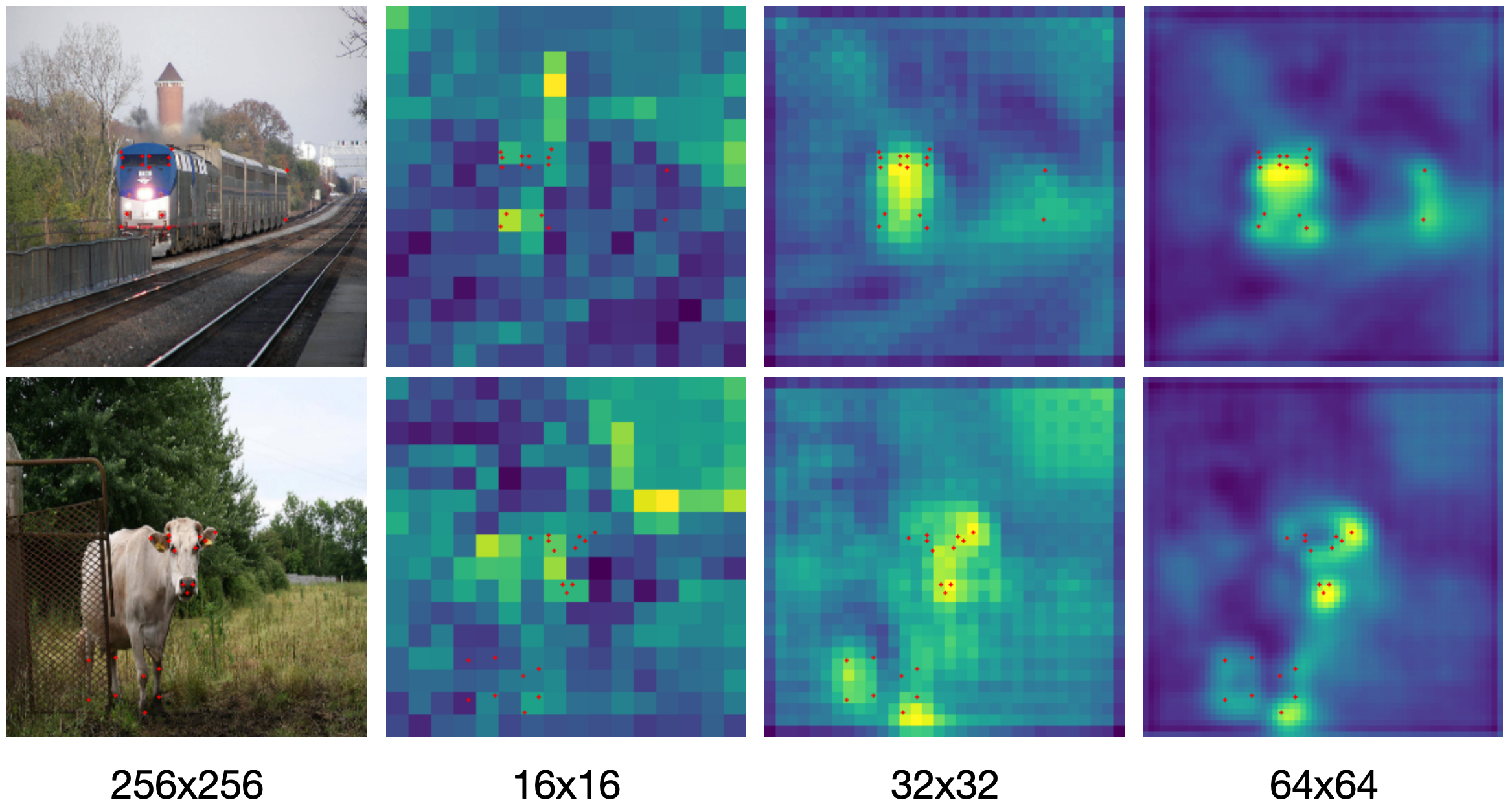}
    \caption{Feature map visualizations at different scales. The red dots represent keypoints.}
    \label{keypoints}
\end{figure*}

As a mainstream semantic correspondence paradigm, 4D-correlation based methods \cite{rocco2018neighbourhood,cho2021cats,kim2022transformatcher} address correspondence inconsistencies through cost aggregation techniques. While these approaches demonstrate computational efficiency, their performance remains constrained by the inherent limitations of 4D decoders in handling higher resolution feature maps. In contrast, recent feature-based methods \cite{li2024sd4match,zhang2024telling} that leverage pre-trained large-scale vision models have achieved remarkable performance by exploiting rich semantic representations. However, this performance gain comes at an expensive computational cost due to their reliance on high-resolution inputs. The current situation reveals a trade-off in  semantic correspondence research: while 4D-correlation methods offer efficient but limited matching capability, feature-based approaches provide superior accuracy but suffer from high computational cost, highlighting the need for architectures that can achieve both high accuracy and efficiency simultaneously.

In this paper, we present SimpleMatch, a surprisingly simple yet effective framework for semantic correspondence that requires only a lightweight upsampling decoder. Our key insight stems from the observation that deep downsampling operations in backbones cause distinct keypoint features to merge irreversibly, fundamentally impairing the model's ability to establish precise correspondences. As illustrated in Figure~\ref{keypoints}, when a $256\times256$ input image is downsampled to a $16\times 16$ feature map, multiple semantically distinct keypoints inevitably collapse into a single feature vector, which may result in incorrect matches in the target image. To address this issue, we introduce an efficient upsampling decoder that restores features to $1/4$ input resolution to separate merged keypoints, and a multi-scale supervised loss that enhances feature discriminability. Thanks to its simplicity and effectiveness, we expect SimpleMatch to become a strong baseline for semantic correspondence.

While deep upsampling improves feature resolution for semantic correspondence, it incurs substantial memory overhead during training. To address this, we propose two complementary strategies: sparse matching and window-based localization. Unlike conventional dense matching approaches that exhaustively compute correspondences between all features, our sparse matching selectively processes only the most relevant source keypoints. For target keypoint localization, we first estimate a coarse search window and then refine the precise position within this restricted region. Collectively, these strategies achieve a 51\% reduction in training memory consumption.

Our contributions can be summarized as follows:

\begin{itemize}

    \item We introduce SimpleMatch, a simple yet effective framework for semantic correspondence that combines a lightweight upsampling decoder with a multi-scale supervised loss to enhance the feature representation.
    
    \item We develop memory-efficient sparse matching and window-based localization techniques that reduce training memory usage by 51\%.
    
    \item Extensive experiments show that our method outperforms state-of-the-art approaches, achieving a PCK@0.1 score of 84.1\% on SPair-71k while using a $3.3\times$ lower input resolution, thereby demonstrating both efficiency and effectiveness.
\end{itemize}

\section{Related Work}

\subsection{Semantic Correspondence}

Semantic correspondence seeks to establish pixel-level correspondences between images depicting instances of the same category. Early approaches, such as SIFT \cite{lowe2004distinctive} and SURF \cite{bay2006surf}, relied on hand-crafted local features to establish correspondences, but were limited in capturing high-level semantics. The advent of Convolutional Neural Networks (CNNs) addresses this issue by enabling end-to-end learning of discriminative features. Rocco \textit{et al.} \cite{rocco2017convolutional} introduced a CNN-based framework combining feature matching and geometric transformation estimation to handle noisy correspondences. Yang \textit{et al.} \cite{yang2017object} proposed an object-aware hierarchical graph model, training the model in an objectness-driven manner. The subsequent progress has been driven by 4D correlation-based methods, which model dense correspondences in a 4D space. NCNet \cite{rocco2018neighbourhood}, the pioneering work in this line, employed 4D convolutions to identify spatially consistent matches by refining 4D correlation maps. Building upon this, ANC-Net \cite{li2020correspondence} introduced non-isotropic 4D convolutions for robust matching, which were used to build an adaptive neighbourhood consensus module. CATs \cite{cho2021cats} and TransforMatcher \cite{kim2022transformatcher} utilize self-attention mechanisms to explore global consensus based on the 4D correlation map. However, 4D space computations incur high computational cost, which hinders their scalability to higher-resolution feature maps, posing a challenge for fine-grained matching. 

Recent progress in semantic correspondence has seen the successful application of large-scale vision models. DIFT \cite{tang2023emergent} demonstrates that Stable Diffusion (SD) \cite{rombach2022high} features exhibit emergent correspondence capabilities through simple nearest neighbor matching. SD+DINO \cite{zhang2024tale} combines SD features with self-supervised DINO representations to achieve more robust matching performance. Subsequently, SD4Match \cite{li2024sd4match} and GeoAware-SC \cite{zhang2024telling} further advanced supervised semantic correspondence by either fine-tuning the backbone or introducing learnable parameters within end-to-end frameworks. While these methods have achieved significant performance improvements, they typically require high input resolutions, such as $512 \times 512$ or $960\times960$ for SD models, which adversely affects computational efficiency and practical applications. In this work, we investigate a more efficient architecture that achieves superior performance while operating at lower input resolutions.

\subsection{Feature Upsampling}

High-resolution features have been widely used in computer vision tasks that demand fine-grained image understanding, including semantic segmentation \cite{strudel2021segmenter,zheng2021rethinking}, object detection \cite{wang2024yolov10, girshick2015fast}, and image generation \cite{rombach2022high}. These applications benefit substantially from high-resolution representations, as they enable the capture of detailed spatial information and precise localization of visual elements. However, high-resolution features remain underutilized in semantic correspondence due to the high memory cost of processing 4D correlation maps. To overcome these limitations, we rethink the conventional pipeline and eliminate the computationally expensive refinement step of 4D score maps, instead using these maps directly for predicting target keypoint coordinates. Additionally, we introduce sparse matching and window-based localization to further reduce the training memory overhead.

\section{Method}

We first investigate a long-standing but overlooked limitation in semantic correspondence, and then introduce our proposed solution, a simple yet effective method to address this issue. Subsequently, we present two complementary strategies: sparse matching and window-based localization, designed to enhance training efficiency. Lastly, we detail the proposed training loss function.

\subsection{A fundamental limitation in semantic correspondence}

Conventional approaches in semantic correspondence tasks have widely adopted low-resolution inputs, such as $256\times256$, combined with $16\times$ downsampling as standard practice \cite{cho2021cats,sun2023correspondence,sun2024pixel}. While this configuration offers computational efficiency, it introduces significant representational challenges. Our analysis reveals that this approach causes distinct keypoints to collapse into a single feature vector, with up to 7 keypoints being fused into a single feature vector in the worst cases.

For quantitative analysis, We performed a detailed statistical analysis of keypoint distributions across training datasets. As detailed in Table~\ref{count}, SPair-71k exhibits high sensitivity to this issue, with 18.8\% of keypoints being affected at $16\times$ downsampling. This effect exhibits strong inverse correlation with feature resolution, decreasing to merely 0.8\% when the resolution is upsampled to $64\times64$. The PF-PASCAL dataset shows greater robustness, with only 4.3\% of keypoints impacted at $16\times$ downsampling, further reducing to 0.2\% when using 1/4 resolution feature maps.

\begin{table}
    \centering
    \begin{tabular}{cccc c}
        \toprule
        Feature map size & 16x16 & 32x32 & 64x64 & total\\
        \midrule
        SPair-71k & 1957 & 524 & 88 &  10408  \\
        PF-PASCAL & 352 & 66 & 14 & 8180 \\
        \bottomrule
    \end{tabular}
    \caption{Numerical statistics of keypoint fusion across feature map resolutions. The input image size is $256\times256$.}
    \label{count}
\end{table}

\begin{table}
    \centering 
    \begin{tabular}{ c | c c c | c c c }
        \toprule
        \multirow{2}{*}{Methods}  & \multicolumn{3}{c}{SPair-71k} & \multicolumn{3}{c}{PF-PASCAL}  \\
         & 0.05 & 0.1 & 0.15 & 0.05 & 0.1 & 0.15  \\
        \midrule
        Ours$_{16\times16}$ & 29.4 & 51.0 & 62.8 & 72.4 & 90.0 & 95.2 \\
        Ours$_{32\times32}$ & 33.9 & 54.0 & 64.5 & 77.7 & 91.2 & 95.1 \\
        Ours$_{64\times64}$ & 43.2 & 59.6 & 67.7 & 79.9 & 91.8 & 95.3 \\
        \bottomrule
    \end{tabular}
    \caption{Performance evaluation of SimpleMatch across multiple feature map resolutions.}
    \label{resolution}
\end{table}

Experimental results across different feature map scales confirm these findings, as shown in Table~\ref{resolution}. SPair-71k displays pronounced resolution sensitivity, exhibiting an 8.6\% performance differential between $16\times16$ and $64\times64$ feature maps. In contrast, PF-PASCAL maintains relatively stable performance across scales, showing only a 1.8\% variation under the same conditions. This divergence suggests not only differences in dataset composition, but also reflects real-world keypoint distribution patterns. Thus, we recommend that future architectural designs take this resolution dependency into account.

\begin{figure*}
    \centering
    \includegraphics[scale=0.79]{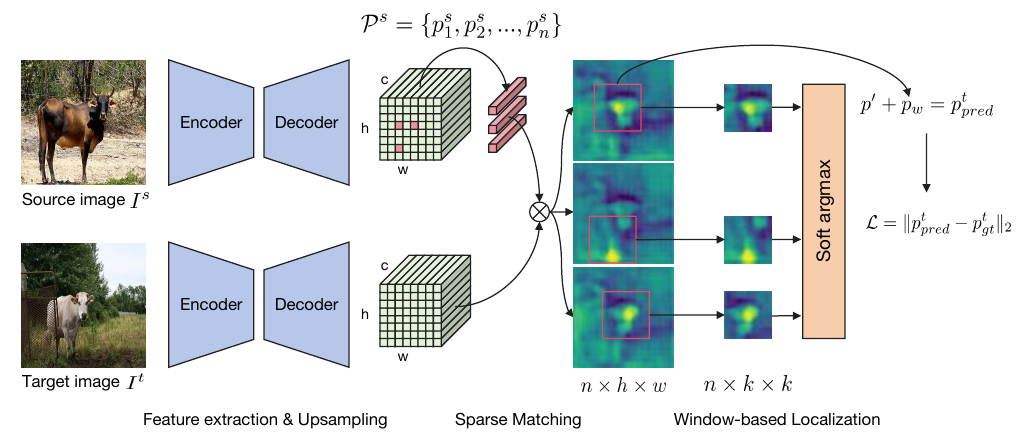}
    \caption{\textbf{Illustration of SimpleMatch structure}. The architecture consists solely of a feature extractor and a lightweight upsampling decoder. After obtaining the source and target feature maps, we perform sparse matching and employ window-based localization to enhance training efficiency.}
    \label{model}
\end{figure*}

\subsection{A simple and strong baseline for semantic correspondence}

Based on the above analysis, we propose SimpleMatch, a simple yet effective framework designed to serve as a new baseline for semantic correspondence. Our proposed architecture, illustrated in Figure~\ref{model}, comprises two main components: a feature encoder to extract image representations and a lightweight decoder for upsampling and refining the feature maps. Unlike existing methods that rely on complex matching modules, SimpleMatch adopts a minimalist design that ensures computational efficiency while achieving superior matching performance.

Given an input image $I \in \mathbf{R}^{H\times W \times 3}$, we first extract its deep features using a shared encoder:
\begin{equation}
    F_{\text{encoder}} = \text{Encoder}(I)
\end{equation}
where the $\text{Encoder}$ can be any standard backbone architecture, such as ResNet, ViT, Swin Transformer, etc.

\begin{figure}[ht]
    \centering
    \includegraphics[width=\columnwidth]{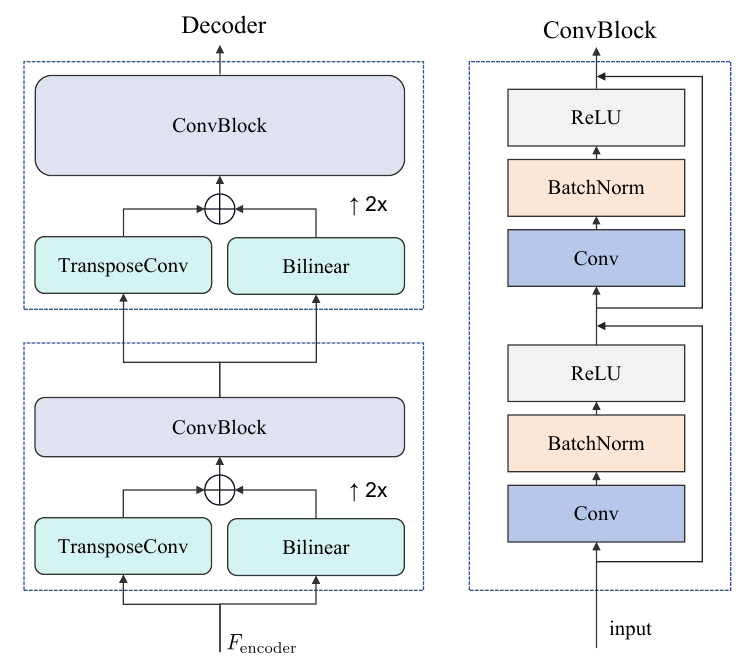}
    \caption{\textbf{Illustration of the lightweight decoder.}}
    \label{decoder}
\end{figure}

The proposed lightweight decoder then progressively upsamples the low-resolution feature maps by a total factor of 4x to recover spatial details, producing a higher-resolution feature representation:
\begin{equation}
    F_{\text{decoder}} = \text{Decoder}(F_{\text{encoder}})
\end{equation}
where $F_{\text{decoder}} \in \mathbf{R}^{\frac{H}{4}\times \frac{W}{4} \times C}$ denotes the high-resolution feature map utilized for subsequent correspondence matching.

As illustrated in Figure~\ref{decoder}, the proposed decoder consists of two identical upsampling modules. Taking the feature $F_{\text{encoder}}$ as input, the module first performs parallel upsampling via transposed convolution and bilinear interpolation, and then fuses the results through summation:
\begin{equation}
    X = \text{TransposeConv}(F_{\text{encoder}}) + \text{Bilinear}(F_{\text{encoder}})
\end{equation}
Notably, when the backbone provides multi-level features (e.g., ResNet), we replace the transposed convolution with skip connections from the corresponding encoder stage to better preserve fine-grained spatial details.

The fused feature $X$ is subsequently refined by a lightweight convolutional block:
\begin{equation}
    X = \text{ConvBlock}(X)
\end{equation}
This ConvBlock consists of two 3x3 convolutional layers with residual connections, offering a good balance between expressiveness and parameter efficiency.

To maintain generality and simplicity, we intentionally avoid incorporating complex components such as attention mechanisms, transformer layers, or iterative refinement modules. Our experiments demonstrate that even with this extremely lightweight decoder design, high-quality correspondence maps can still be produced, leading to state-of-the-art matching performance across multiple benchmarks. This design not only enhances computational efficiency but also establishes SimpleMatch as a robust and efficient baseline for future research.


\subsection{Sparse Matching and Window-based Localization}

While high-resolution feature maps enhance matching precision, they inevitably incur significant memory and computational overhead due to the exhaustive matching across all possible feature pairs in conventional methods. To address this, we propose a combination of sparse matching and window-based localization. Together, these methods significantly reduce training costs while maintaining high matching accuracy.

\subsubsection{Sparse Matching}

Given the source feature map $F_s \in \mathbb{R}^{h \times w \times c}$ and the target feature map $F_t \in \mathbb{R}^{h \times w \times c}$, together with a sparse set of source keypoint coordinates:

\begin{equation}
    \mathcal{P}^s = \{p_1^s, p_2^s, \dots, p_n^s\}
\end{equation}
where $n \ll H \times W$ and each $p_i^s \in \mathbb{R}^2$ is typically expressed in the feature-map coordinate system, we first extract the feature representation at each keypoint location using coordinate index:

\begin{equation}
    \mathcal{F} = \bigl\{ F_s(p_i^s) \mid i = 1, \dots, n \bigr\}
\end{equation}

These $n$ feature vectors are then stacked row-wise to form the keypoint feature matrix

\begin{equation}
F_p \in \mathbb{R}^{n \times C}, \qquad F_p[i,:] = \bigl(F_s(p_i^s)\bigr)^\top
\end{equation}

To compare each source keypoint against the entire target image in an efficient manner, we reshape the target feature map by flattening its spatial dimensions:

\begin{equation}
\widetilde{F}_t \in \mathbb{R}^{M \times C}, \quad M = H \times W
\end{equation}

Finally, we compute the matrix of cosine similarities between the source keypoint descriptors and all spatial locations in the target feature map:

\begin{equation}
\mathcal{S} = \frac{F_p \,\widetilde{F}_t^\top}{\|F_p\|_F \|\widetilde{F}_t\|_F} \in \mathbb{R}^{n \times M}
\label{eq:cosine-similarity}
\end{equation}
where $\|\cdot\|_F$ denotes the Frobenius norm. Each entry $\mathcal{S}_{i,j}$ represents the cosine similarity between the $i$-th source keypoint and the feature at the $j$-th spatial position in the target map.

\subsubsection{Window-based Localization}

For each keypoint $i$, the coarse location in the target feature space is first obtained by identifying the position with the maximum similarity score:

\begin{equation}
    p^i = \arg\max_{j=1,\dots,HW} \mathcal{S}[i,j]
    \label{eq:coarse-location}
\end{equation}

The similarity matrix $\mathcal{S}$ is subsequently reshaped into a per-keypoint spatial map $S \in \mathbb{R}^{n \times H \times W}$. Around each coarse location $p^i = (p_x^i, p_y^i)$, a local window of size $k \times k$ is extracted:
\begin{equation}
\begin{aligned}
    S_i = S\ [i, & \ p_y^i - \frac{k-1}{2} \!:\! p_y^i + \frac{k-1}{2} + 1, \\
     &\ p_x^i - \frac{k-1}{2} \!:\! p_x^i + \frac{k-1}{2} + 1 ]
    \label{eq:local-window}
\end{aligned}
\end{equation}
where $k$ is typically an odd integer. By restricting subsequent refinement to this small $k \times k$ region, the matching complexity is reduced from $\mathcal{O}(HW)$ to $\mathcal{O}(k^2)$ per keypoint. Within the focused window, a fine-grained localization method is applied to predict a precise relative offset $(x', y')$.

The final predicted target keypoint coordinate is recovered via

\begin{equation}
    p_{\text{pred}}^i = \left( p_x^i - \frac{k-1}{2} + x', \\
    \ p_y^i - \frac{k-1}{2} + y' \right)
    \label{eq:final-prediction}
\end{equation}

This coarse-to-fine strategy achieves an effective trade-off between computational efficiency and localization precision for correspondence in high-resolution images.

\subsection{Training Loss}

Existing approaches \cite{cho2021cats,sun2023correspondence,sun2024pixel} typically employ flow estimation as their training objective, where models predict offset vectors from source to target keypoints. By contrast, we propose a more straightforward formulation that directly regresses the absolute coordinates of target keypoints. This simplification eliminates the need for explicit offset computation while maintaining performance.

Mathematically, given ground truth target keypoint coordinates $\mathcal{P}^t_{gt} = \{p_1^t, p_2^t, ..., p_n^t\}$ and predicted positions $\mathcal{P}^t_{pred}$, we define our base loss function as the L2 distance between predicted and ground truth positions:

\begin{equation}
    \mathcal{L} = \| \mathcal{P}^t_{pred} - \mathcal{P}^t_{gt}  \|_2
\end{equation}

To enhance the keypoint representation, we introduce a multi-scale supervision strategy in our lightweight decoder. Our network generates predictions at three different resolutions: 1/16, 1/8 and 1/4. The composite loss combines supervision across all scales:

\begin{equation}
    \mathcal{L} = \mathcal{L}_{1/4} + \mathcal{L}_{1/8} + \mathcal{L}_{1/16}
\end{equation}

\section{Experiments}
\subsection{Experimental Settings}

\noindent\textbf{Datasets.} We evaluate our method on four public benchmarks: SPair-71k \cite{min2019spair}, PF-PASCAL \cite{ham2017proposal}, PF-WILLOW \cite{ham2016proposal}, and AP-10K \cite{zhang2024telling}. SPair-71k is a challenging dataset comprising 70,958 image pairs across 18 categories with large intra-class variations. PF-PASCAL includes 1,351 image pairs spanning 20 categories, while PF-WILLOW consists of 900 test samples distributed across 4 categories. AP-10K, built upon existing animal pose estimation dataset, contains 10015 images covering 23 families and 54 species of animals.

\begin{table*}[!htb]
    \centering
    \resizebox{1\textwidth}{!}{
    \begin{tabular}{l | c  c | c c | c c c| c c}
        \toprule
        \multirow{3}{*}{Methods} & \multirow{3}{*}{Backbone} & \multirow{3}{*}{Resolution} & \multicolumn{2}{c|}{SPair-71k} & \multicolumn{3}{c}{PF-PASCAL} & \multirow{3}{*}{\makecell{Throughput\\(image/s)}} & \multirow{3}{*}{\makecell{Memory\\(GB)}} \\
        & & &  \multicolumn{2}{c|}{$\alpha$: bbox} & \multicolumn{3}{c}{$\alpha$: img} \\
        & & &  0.05 & 0.1 & 0.05 & 0.1 & 0.15 \\
        \midrule

        DHPF~\cite{min2020learning} & ResNet101 & $240\times240$ & - & 37.3 & 75.7 & 90.7 & 95.0 & 44 & 3.5 \\
        CHM~\cite{min2021convolutional} & ResNet101 & $240\times240$ & 22.7 & 46.3 &  80.1 & 91.6 & 94.9 & 41 & 2.6 \\
        CATs~\cite{cho2021cats} & ResNet101 & $256\times256$  & 27.7 & 49.9 & 75.4 & \underline{92.6} & \underline{96.4} & 92 & 3.0 \\
        MMNet-FCN~\cite{zhao2021multi} & ResNet101 & $224\times320$  & \underline{33.3} & 50.4 & \textbf{81.1} & 91.6 & {95.9} & 15 & 6.9 \\
        TransforMatcher~\cite{kim2022transformatcher} & ResNet101 & $240\times240$  & - & {53.7} & \underline{80.8} & {91.8} & - & 133 & 2.7  \\

        KBCNet~\cite{jin2025kbcnet} & ResNet101 & $256\times256$ & - & \underline{59.1} &78.1& \textbf{93.8} & \underline{97.2} & - & -  \\

        \textbf{SimpleMatch (ours)} & ResNet101 & $256\times256$  & \textbf{43.2} & \textbf{59.6} & 79.9& {91.8} & 95.3 & 98 & 2.7 \\
        \midrule 

        CATs~\cite{cho2021cats} & iBOT & $256\times256$  & 30.7 & 55.2 & 77.8 & 93.1 & 96.8 & - & -  \\
        TransforMatcher~\cite{kim2022transformatcher} & iBOT & $256\times256$  & 33.1 & 57.9 & 77.3 & 93.3 & 96.6 & - & -  \\
        ACTR~\cite{sun2023correspondence} & iBOT & $256\times256$  & 42.0 & 62.1 & 81.2 & 94.0 & 97.0 & 44 & 4.9  \\
        LPMFlow~\cite{sun2024pixel} & iBOT & $256\times256$  & \underline{46.7} & \underline{65.6} & \underline{82.4} & \underline{94.3} & \underline{97.2} & 36 & 5.6  \\
        \textbf{SimpleMatch (ours)} & iBOT & $256\times256$  & \textbf{55.1} & \textbf{72.2} & \textbf{84.6} & \textbf{95.5} & \textbf{98.0} & 73 & 3.1  \\

        \midrule
        DHF~\cite{luo2024diffusion} & SD & $512\times512$  &  50.2 & 64.9 & 78.0 & 90.4 & 94.1 & $<$ 1 & 12.5 \\
        SD4Match~\cite{li2024sd4match} & SD & $768\times768$  & 59.5 & 75.5 & 84.4 & {95.2} & {97.5} & 1 & 12.5 \\
        GeoAware-SC~\cite{zhang2024telling} & SD+DINOv2 & $960^2\times840^2$  & \underline{72.8} & {83.2} & {85.3} & {95.0} & {97.4} & $<$ 1& 11.8  \\
        \textbf{SimpleMatch (ours)} & DINOv2 & $252\times252$  & 69.3 & \underline{84.2} & \underline{87.7} & \textbf{96.5} & \textbf{98.4} & 65 & 2.8 \\
        \textbf{SimpleMatch (ours)} & DINOv2 & $448\times448$  & \textbf{79.2} & \textbf{88.2} & \textbf{88.3} & \underline{95.9} & \underline{97.6} & 26 & 3.0 \\

        \bottomrule
        
    \end{tabular}

    }
    \caption{\textbf{Evaluation on SPair-71k and PF-PASCAL}. We present the performance evaluation of SimpleMatch with different backbone architectures, including inference throughput and memory consumption measurements. All models are tested on NVidia RTX 4090 GPU with 24GB memory. The best results are in \textbf{bold}, and the second-best results are \underline{underlined}. Higher PCK is better.}
    \label{tab:sota}
    
\end{table*}

\noindent\textbf{Evaluation metrics.} Following standard evaluation protocols, we employ the percentage of correct keypoints (PCK@$\alpha$) metric, which measures the proportion of estimated keypoints falling within a specified threshold distance from their ground-truth positions. Given a set of predicted keypoints $\mathcal{P}^{pred}$ and ground-truth keypoints $\mathcal{P}^{gt}$, PCK is calculated as $PCK = \frac{1}{n} \sum_{i=1}^n \mathbb{I}(d(\mathcal{P}^{pred}_i, \mathcal{P}^{gt}_i) < \alpha \cdot \max(H, W))$, where $d(\cdot)$ denotes the Euclidean distance, $H$ and $W$ represent either the image dimensions or the object's bounding box size, and $\alpha$ is the threshold. Here, $\mathbb{I}$ is the indicator function, which evaluates to 1 if the condition is met and 0 otherwise.

\noindent\textbf{Implementation Details.} For feature extraction, we evaluate three distinct backbone architectures: ResNet101, iBOT, and DINOv2. To enhance generalization, we apply data augmentation techniques, including random cropping, random rotation, and appearance transformations such as color jitter, posterization, and sharpening. The window size is set to 45 for both iBOT and DINOv2 and to 30 for ResNet101. For SPair-71k and AP-10k, we employ an initial learning rate of 6e-4, whereas for PF-PASCAL we use a learning rate of 1e-4. Additionally, we apply a step-based learning rate scheduler for decaying the learning rate. The decay step size depends on the total training iterations, and we set the step size to 800, 2000, 200 for SPair-71k, AP-10K and PF-PASCAL, respectively. The decay factor is set to 0.95. The multi-scale loss is employed for ResNet101 to enhance the representations.  We train the models for 10, 10 and 100 epochs on SPair-71k, AP-10K and PF-PASCAL, respectively. All experiments are implemented in PyTorch and trained on a single NVIDIA GeForce RTX 4090 GPU.

\subsection{Comparison with State-of-The-Art}

\begin{figure}
    \centering
    \includegraphics[width=\columnwidth]{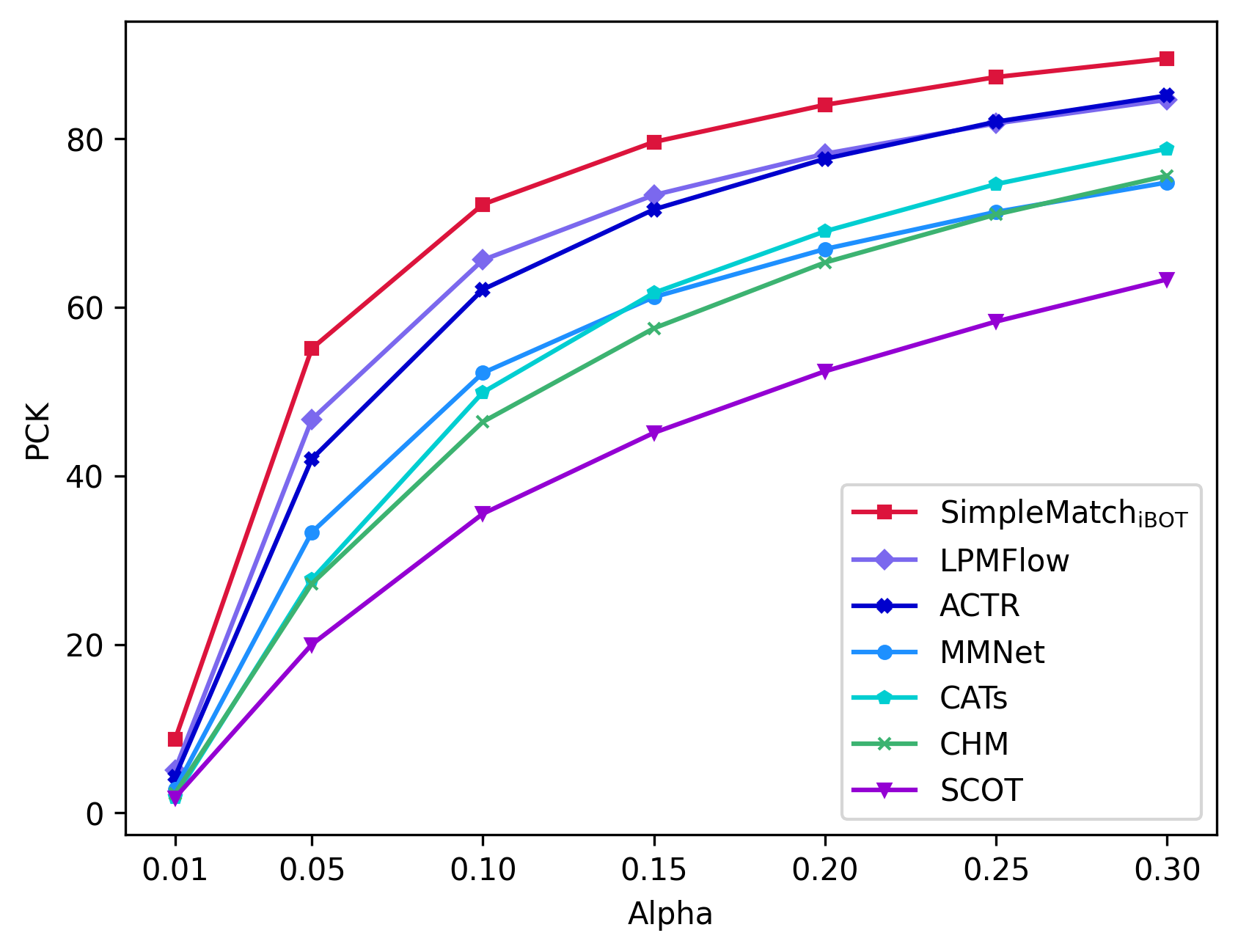}
    \caption{The PCK@$\alpha$ curves of SimpleMatch and previous methods on SPair-71k dataset.}
    \label{fig:alpha}
\end{figure}

\begin{table*}[t]
    \centering
    \resizebox{1\textwidth}{!}{
    \begin{tabular}{lcccccccccccccccccccl}
    \toprule
    {Method}  & Aero & Bike & Bird & Boat &Bottle&Bus&Car&Cat&Chair& Cow& Dog& Horse& Motor&Person& Plant& Sheep & Train & TV & \textbf{All}\\  
    \midrule
    SCOT~\cite{liu2020semantic} & 34.9  & 20.7 & 63.8 & 21.1 & 43.5  & 27.3 & 21.3 & 63.1 & 20.0  & 42.9 & 42.5 & 31.1  & 29.8  & 35.0  & 27.7  & 24.4  & 48.4  & 40.8 & 35.6 \\
    DHPF~\cite{min2020learning}       & 38.4  & 23.8 & 68.3 & 18.9 & 42.6  & 27.9 & 20.1 & 61.6 & 22.0  & 46.9 & 46.1 & 33.5  & 27.6  & 40.1  & 27.6  & 28.1  & 49.5  & 46.5 & 37.3 \\
    CHMNet~\cite{min2021convolutional} & 49.1  & 33.6 & 64.5 & {32.7} & 44.6  & 47.5 & 43.5 & 57.8 & 21.0  & 61.3 & 54.6 & 43.8  & 35.1  & {43.7}  & 38.1  & {33.5}  & 70.6  & 55.9 & 46.3 \\
    MMNet~\cite{zhao2021multi}      & 43.5  & 27.0 & 62.4 & 27.3 & 40.1  & 50.1 & 37.5 & 60.0 & 21.0  & 56.3 & 50.3 & 41.3  & 30.9  & 19.2  & 30.1  & 33.2  & 64.2  & 43.6 & 40.9 \\
    
    CATs~\cite{cho2021cats} &  52.0 & 34.7 & 72.2 & 34.3 & 49.9 & 57.5 & 43.6 & 66.5 & 24.4 & 63.2 & 56.5 & 52.0 & 42.6 & 41.7 & {43.0} & 33.6 & 72.6 & 58.0 & 49.9 \\
    
    MMNet-FCN~\cite{zhao2021multi} & \underline{57.0} & 37.0 & 66.3 & 35.8  & 48.9 & 63.4 & 50.1 & 66.6 & \textbf{31.2} & {70.6} & {56.8} & 53.2 & 41.7 & 35.7 & 40.8 & {36.7} & \underline{81.5} & 68.0 & 52.2 \\
    
    TransforMatcher~\cite{kim2022transformatcher} & \textbf{59.2} & {39.3} & {73.0} & \underline{41.2} & {52.5} & \underline{66.3} & \underline{55.4} & {67.1} & {26.1} &  67.1 & 56.6 & {53.2} & {45.0} & {39.9} & 42.1 & 35.3 & 75.2 & {68.6} & {53.7} \\

    KBCNet~\cite{jin2025kbcnet} & 55.6 & \textbf{48.2} & \underline{77.1} & {38.0} & \underline{56.4} & 62.3 & 49.7 & \underline{69.2} & \underline{31.1} & \underline{73.7} & \underline{64.7} & \textbf{63.0} & \textbf{52.7} & \textbf{58.1} & \textbf{59.5} & \underline{46.3} & 81.3 & \underline{74.2} & \underline{59.1} \\

    \textbf{SimpleMatch}$_\mathrm{ResNet101}$ & 53.3 & \underline{46.0} & \textbf{78.2} & \textbf{42.8} & \textbf{62.0} & \textbf{69.1} & \textbf{62.3} & \textbf{73.2} & 25.3 & \textbf{75.6} & \textbf{68.1} & \underline{59.0} & \underline{50.7} & \underline{46.9} & \underline{47.9} & \textbf{47.8} & \textbf{86.3} & \textbf{79.4} & \textbf{59.6} \\

    \midrule

    ACTR~\cite{sun2023correspondence} & 65.0 & 48.5 & 82.3 & \underline{50.4} & 55.9 & 65.3 & 63.1 & 72.8 & 35.9 & 74.0 & 70.3 & 68.8 & 58.6 & 57.1 & 46.8 & 49.5 & 84.4 & 73.3 & 62.1 \\
    LPMFlow~\cite{sun2024pixel} & \textbf{71.4} & \underline{54.8} & \textbf{83.2} & 50.3 & \underline{56.7} & \underline{75.4} & \underline{68.9} & \underline{79.3} & \underline{41.0} & \underline{78.3} & \underline{74.1} & \underline{73.6} & \underline{58.7} & \underline{57.1} & \underline{48.6} & \underline{54.8} & \underline{87.5} & {74.6} & \underline{65.6} \\

    \textbf{SimpleMatch}$_\mathrm{iBOT}$ & \underline{70.1} & \textbf{58.0} & \textbf{85.8} & \textbf{59.4} & \textbf{77.7} & \textbf{78.3} & \textbf{74.8} & \textbf{79.7} & \textbf{48.2} & \textbf{80.4} & \textbf{81.9} & \textbf{74.8} & \textbf{66.9} & \textbf{69.5} & \textbf{57.7} & \textbf{58.1} & \textbf{89.4} & \textbf{90.8} & \textbf{72.2} \\

    \midrule

    DHF~\cite{luo2024diffusion}  & 74.0 & 61.0 & 87.2 & 40.7 & 47.8 & 70.0 & 74.4 & 80.9 & 38.5 & 76.1 & 60.9 & 66.8 & 66.6 & 70.3 & 58.0 & 54.3 & 87.4 & 60.3 & 64.9 \\
    
    SD+DINO (S)~\cite{zhang2024tale}  & {81.2} & {66.9} & {91.6} & {61.4} & {57.4} & {85.3} & {83.1} & {90.8} & {54.5} & {88.5} & {75.1} & {80.2} & {71.9} & {77.9} & {60.7} & {68.9} & {92.4} & {65.8} & {74.6} \\

    SD4Match~\cite{li2024sd4match} & 75.3& 67.4 &85.7& 64.7& 62.9 &86.6& 76.5 &82.6& 64.8& 86.7& 73.0& 78.9 &70.9 &78.3 &66.8& 64.8& 91.5& \underline{86.6} & 75.5 \\

    GeoAware-SC~\cite{zhang2024telling} & \underline{87.6} & \underline{74.1} & \underline{95.5} & \underline{70.1} & \underline{66.7} & \underline{92.0} & \underline{87.4} & \underline{91.4} & \underline{68.0} & \underline{93.2} & \underline{85.5} & \underline{84.7} & \underline{79.9} & \underline{87.8} & \textbf{79.9} & \underline{78.9} & \textbf{96.9} & {84.8} & \underline{83.2} \\


    \textbf{SimpleMatch}$_\mathrm{DINOV2}$ & \textbf{94.2} & \textbf{76.8} & \textbf{96.9} & \textbf{77.2} & \textbf{85.0} & \textbf{95.5} & \textbf{89.6} & \textbf{94.1} & \textbf{81.9} & \textbf{93.6} & \textbf{93.9} & \textbf{89.9} & \textbf{82.4} & \textbf{90.6} & \underline{77.8} & \textbf{80.2} & \underline{96.1} & \textbf{96.9} & \textbf{88.2} \\

    \bottomrule
    \end{tabular}
    } 
    \caption{\textbf{Per-category results on SPair-71k}. We report the detailed per-category results on SPair-71k dataset using PCK@0.1 metric. The best results are in \textbf{bold}, and the second-best results are \underline{underlined}.}
    \label{tab:category}
    
\end{table*}

\begin{table*}[ht]
    \centering
    \begin{tabular}{l | ccc | ccc | cc}
        \toprule
        \multirow{3}{*}{Methods} & \multicolumn{3}{c|}{PF-PASCAL $\rightarrow$ SPair-71k} & \multicolumn{3}{c|}{SPair-71k $\rightarrow$ PF-PASCAL} & \multicolumn{2}{c}{PF-PASCAL $\rightarrow$ PF-WILLOW} \\
        
        & \multicolumn{3}{c|}{$\alpha$: bbox} & \multicolumn{3}{c|}{$\alpha$: img} & $\alpha$: bbox & $\alpha$: kps \\
        & \makebox[0.05\textwidth][c]{0.05} & \makebox[0.05\textwidth][c]{0.1} & \makebox[0.05\textwidth][c]{0.15} & \makebox[0.05\textwidth][c]{0.05} & \makebox[0.05\textwidth][c]{0.1} & \makebox[0.05\textwidth][c]{0.15} & \makebox[0.1\textwidth][c]{0.1} & \makebox[0.1\textwidth][c]{0.1} \\
        \midrule
        ACTR~\cite{sun2023correspondence} & 14.4 & 28.1 & 37.6 & 56.3 & 74.6 & 83.0 & 87.2 & 79.9 \\
        LPMFlow~\cite{sun2024pixel} & 17.6 & 31.6 & 40.8 & 57.3 & 73.2 & 80.7 & 87.6 & 81.0 \\
        \textbf{SimpleMatch}$^\dagger$ & \textbf{24.1} & \textbf{40.6} & \textbf{50.7} & \textbf{66.3} & \textbf{81.9} & \textbf{89.1} & \textbf{89.5} & \textbf{83.6} \\
        \bottomrule
    \end{tabular}
    \caption{\textbf{Transfer learning evaluation}. We report the transfer learning performance of the model that was trained on one dataset and tested on another. The best results are in \textbf{bold}. $\dagger$ denotes using iBOT as the backbone.}
    \label{tab:robust}
\end{table*}

We compare our proposed method with state-of-the-art (SOTA) approaches on both SPair-71k and PF-PASCAL datasets, as presented in Table \ref{tab:sota}. To ensure a comprehensive and fair comparison, we assess three distinct backbones and specify the corresponding input resolutions. The results indicate that despite its simplicity, SimpleMatch achieves notable improvements on SPair-71k across all backbones. Specifically, SimpleMatch$_{\mathrm{ResNet101}}$ outperforms TransforMatcher by 5.9\%, while SimpleMatch$_{\mathrm{iBOT}}$ surpasses LPMFlow by 6.6\%. These gains can be attributed to the effective handling of adjacent keypoint feature fusion in our method. In contrast, on PF-PASCAL dataset, SimpleMatch does not yield comparable performance improvements. This is primarily due to the distinct characteristics of PF-PASCAL, where feature fusion among keypoints is less challenging. Furthermore, as illustrated in Figure~\ref{fig:alpha}, SimpleMatch demonstrates consistent performance gains over previous methods across various threshold values $\alpha$. Quantitative analysis confirms that SimpleMatch maintains superior accuracy and robustness across multiple evaluation thresholds, highlighting its overall effectiveness.

When compared to recent large-scale model-based methods, our proposed method not only exhibits superior performance but also demonstrates notable inference efficiency. Our proposed SimpleMatch$_\mathrm{DINOv2}$ obtains 84.2\% PCK@0.1 on SPair-71k and 96.5\% PCK@0.1 on PF-PASCAL, thereby outperforming GeoAware-SC by 1.0\% and 1.5\%, respectively. It is worth noting that this performance is attained at a resolution of $252\times252$, which is $3.3\times$ lower than that of the SOTA method. Furthermore, SimpleMatch can process up to 65 images per second with a memory footprint of 2.8 GB. In contrast, previous SD-based methods typically achieve fewer than 1 image per second and require approximately 12 GB of memory. Notably, when the input image size is extended to $448\times448$, SimpleMatch demonstrates an additional 4\% improvement on SPair-71k. These experiments underscore the effectiveness and efficiency of our proposed method.

We further analyze the category-wise performance on SPair-71k, with detailed results summarized in Table~\ref{tab:category}. The experimental findings indicate that SimpleMatch achieves SOTA performance across most categories, demonstrating substantial improvements. For instance, in the "person" category, SimpleMatch$_\mathrm{iBOT}$ exhibits a 12.4\% improvement over LPMFlow, underscoring the superiority of our approach.

To evaluate the generalization ability and robustness of our method, we conduct transfer learning experiments where models trained on one dataset are tested on another. As shown in Table~\ref{tab:robust}, our approach yields significant improvements when compared with LPMFlow and ACTR. Specifically, SimpleMatch outperforms LPMFlow by 9.0\% in transfer learning from PF-PASCAL to SPair-71k and by 8.7\% from SPair-71k to PF-PASCAL. Additional robustness evaluations on AP-10K, presented in Table~\ref{tab:ap10k}, show consistent superiority across intra-species (+6.9\%), cross-species (+8.9\%), and cross-family (+6.7\%) splits using the PCK@0.01 metric. These results confirm the precise localization capabilities of our proposed method.

\begin{table*}[h]
    \centering
    \begin{tabular}{l ccc|ccc|ccc}
    \toprule
    &\multicolumn{3}{c}{AP-10K-I.S.} &\multicolumn{3}{c}{AP-10K-C.S.} &\multicolumn{3}{c}{AP-10K-C.F.}\\
    \cmidrule(l){2-4} \cmidrule(l){5-7} \cmidrule(l){8-10}
    
    {Method} & 0.01 & 0.05 & 0.10& 0.01 & 0.05 & 0.10& 0.01 & 0.05 & 0.10 \\  
    \midrule

    DHF~\cite{luo2024diffusion}  &8.0&45.8&62.7&6.8&42.4&60.0&5.0&32.7&47.8  \\
    SD+DINO(S)~\cite{zhang2024tale}  &{9.9}&{57.0}&{77.0}&{8.8}&{53.9}&{74.0}&{6.9}&{46.2}&{65.8}  \\
    GeoAware-SC~\cite{zhang2024telling} &  \underline{{23.2}} & \underline{{73.2}} & \underline{{87.7}} & \underline{{21.7}} & \underline{{70.3}} & \underline{{85.9}} & \underline{{18.3}} & \underline{{63.2}} & \underline{{78.5}} \\
    \textbf{SimpleMatch}$^\dagger$ & \textbf{30.1} & \textbf{75.7} & \textbf{88.2} & \textbf{30.6} & \textbf{75.8} & \textbf{88.0} & \textbf{25.0} & \textbf{68.6 }& \textbf{81.3} \\
    
    \bottomrule
    \end{tabular}
    \caption{Evaluation results on the AP-10K dataset across intra-species (I.S.), cross-species (C.S.), and cross-family (C.F.) test scenarios. $\dagger$: We train SimpleMatch with the input size of $448\times448$. The best results are in \textbf{bold} and the second best results are \underline{underlined}.}
    \label{tab:ap10k}
\end{table*}

\subsection{Ablation Study}

\textbf{Ablation on window size.} We evaluate the impact of window size on matching performance and present the results in Table~\ref{tab:window}. The window size hyperparameter reflects a trade-off between contextual scope and localization precision in coordinate estimation. The $45\times45$ window achieves the best performance in terms of PCK@0.1 and PCK@0.15 metrics, whereas the $15\times15$ window configuration excels in fine-grained localization. Performance degrades at both extremes: smaller windows ($9\times9$) suffer from limited contextual information, while larger windows ($60\times60$) tend to incorporate noise from irrelevant background regions. Based on these observations and considering common evaluation practices, we adopt $45\times45$ as the default window size for SimpleMatch.  

\begin{table}
    \centering
    \begin{tabular}{c|ccc}
        \toprule
        \multirow{2}{*}{Window size} & \multicolumn{3}{c}{SPair-71k}  \\
        & 0.05 & 0.1 & 0.15 \\
        \midrule
        9 & 55.4 & 69.4 & 85.1 \\
        15 & \textbf{56.4} & 70.6 & 76.3 \\
        30 & 55.3 & 71.5 & 78.2 \\
        45 & 55.1 & \textbf{72.2} & \textbf{79.6} \\
        60 & 52.8 & 70.6 & 78.2 \\
        \bottomrule
    \end{tabular}
    \caption{Ablations on window size.}
    \label{tab:window}
\end{table}


\textbf{Ablation on multi-scale loss.} We conduct ablation studies to evaluate the effectiveness of our proposed multi-scale loss, as shown in Table~\ref{tab:loss}. The multi-scale supervision is introduced to mitigate feature quality degradation in the decoder by enforcing supervised learning across multiple spatial resolutions. Experimental results indicate a 1.4\% drop in performance on PCK@0.1 when the multi-scale loss is omitted. This performance decline validate the importance of multi-level supervision.

\begin{table}
    \centering
    \begin{tabular}{l|ccc}
        \toprule
        \multirow{2}{*}{Loss function} & \multicolumn{3}{c}{SPair-71k}  \\
        & 0.05 & 0.1 & 0.15  \\
        \midrule
        Baseline & 43.2 & 59.6 & 67.7 \\
        w/o multi-scale loss & 42.8 & 58.2 & 65.6 \\
        \bottomrule
    \end{tabular}
    \caption{Ablations on multi-scale loss.}
    \label{tab:loss}
\end{table}

\textbf{Ablation on feature map size.} As illustrated in Table~\ref{tab:feat}, we conduct an analysis of models trained with different input resolutions while maintaining the consistent feature map size. Our observations reveal that the model performance is more related to the feature map resolution rather than simply to the input resolution. For instance, when the feature map size is set to $64\times64$ or $32\times32$, the performance gaps 
\begin{table}[ht]
    \centering
    \resizebox{1.0\linewidth}{!}{
    \begin{tabular}{c|cc}
        \toprule
        \multirow{2}{*}{Feature map size} & \multicolumn{2}{c}{SPair-71k (PCK@0.1)}  \\
        &  SimpleMatch$_{256\times256}$ & SimpleMatch$_{512 \times 512}$ \\
        \midrule
        128x128 & - & 73.9 \\
        64x64 & 72.2 & 72.9  \\
        32x32 & 68.6 & 68.2 \\
        16x16 & 59.7 & - \\
        \bottomrule
    \end{tabular}
    }
    \caption{Ablations on feature map size.}
    \label{tab:feat}
\end{table}
\begin{table}[ht]
    \centering
    \begin{tabular}{c|ccc}
        \toprule
        \multirow{2}{*}{Input resolution} & \multicolumn{3}{c}{SPair-71k}  \\
        & 0.05 & 0.1 & 0.15  \\
        \midrule
        $256\times256$ & 55.1 & 72.2 & 79.6 \\
        320x320 & 59.2 & 74.3 & \textbf{80.7} \\
        384x484 & 60.9 & 74.5 & 80.6 \\
        448x448 & \textbf{62.2} & \textbf{74.7} & 80.3 \\
        512x512 & 61.5 & 73.9 & 79.3 \\
        \bottomrule
    \end{tabular}
    \caption{Ablations on input resolution.}
    \label{tab:resolution}
\end{table}
are only 0.7\% and 0.4\%, respectively.  However, increasing the feature map resolution from $32\times32$ to $64\times64$ leads a more pronounced performance gap of 3.6\%. These findings suggest that enhancing feature map resolution, rather than merely increasing the input resolution, constitutes a straightforward and effective strategy for approaching the performance levels achieved by higher-resolution inputs.

\begin{figure*}
    \centering
    \includegraphics[width=0.9\textwidth]{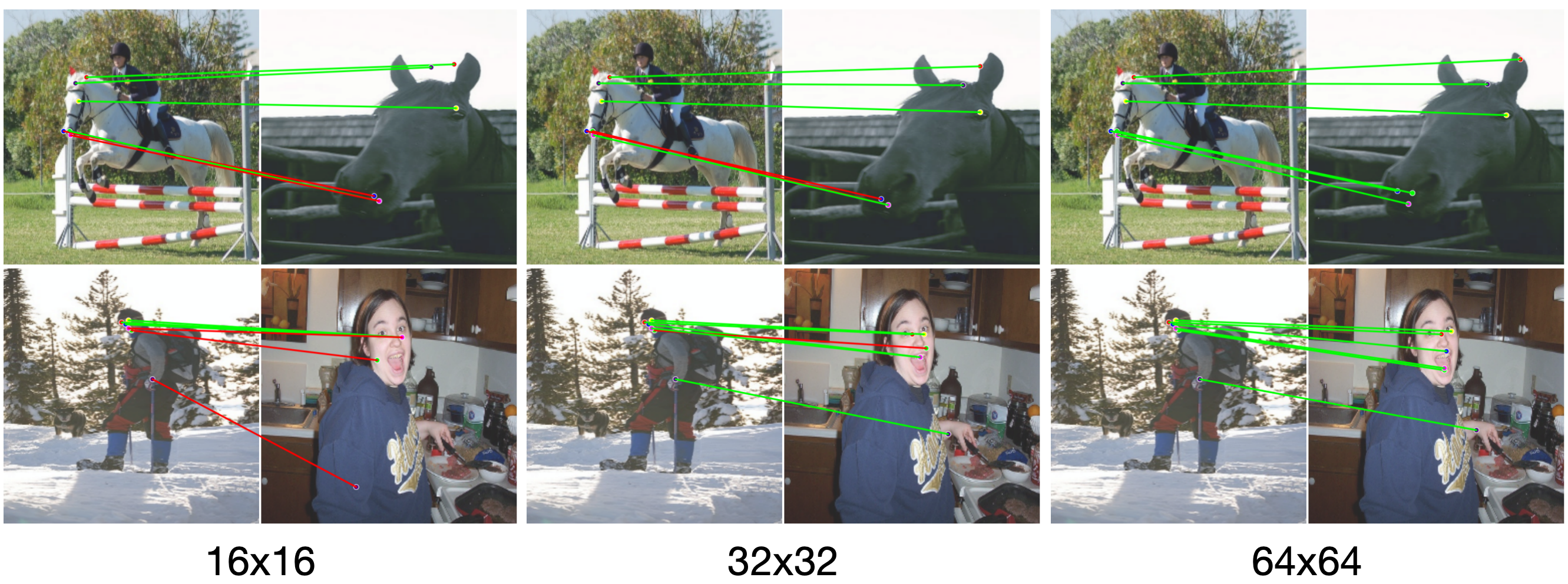}
    \caption{Visualization of semantic correspondence using different feature map resolutions. From left to right, we show the semantic correspondence performance at 16x16, 32x32 and 64x64 resolutions. Green lines indicate correct matches, while red lines denote incorrect matches. The results are evaluated on PCK@0.1.}
    \label{fig:match}
\end{figure*}
\begin{figure*}
    \centering
    \includegraphics[width=0.9\textwidth]{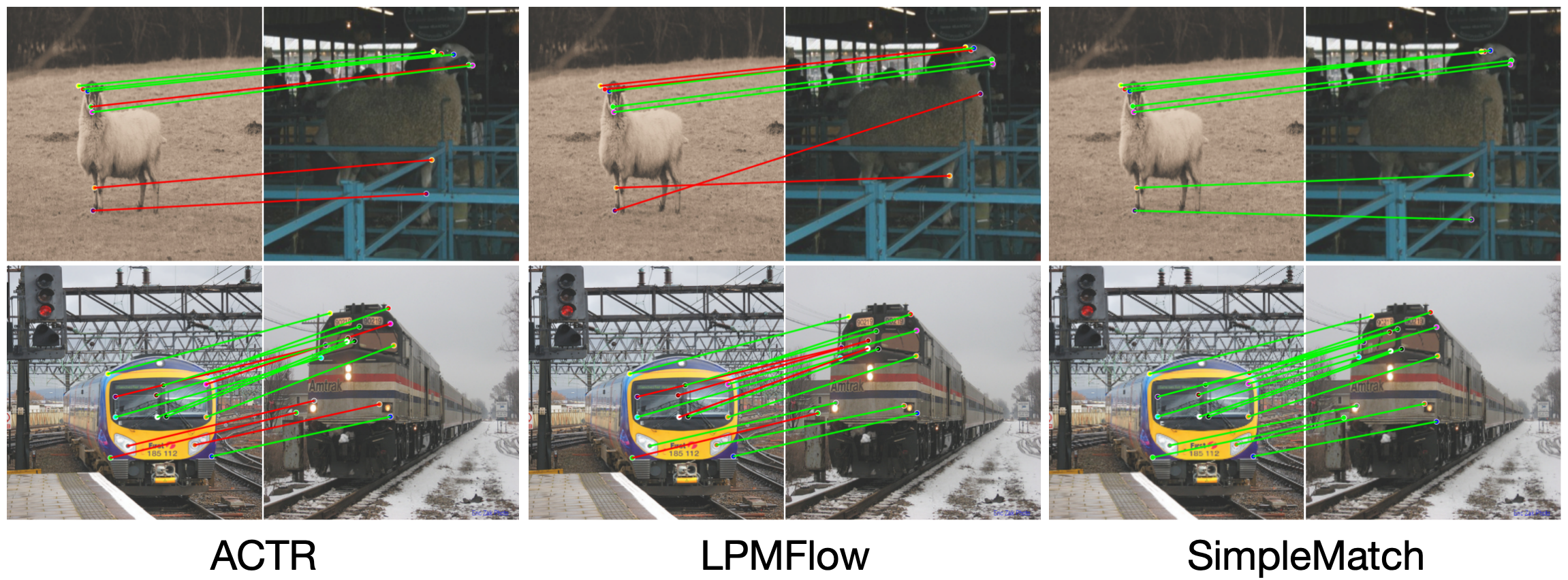}
    \caption{Visualization of semantic correspondence compared with SOTA methods. We report the visual correspondence performance of ACTR, LPMFlow and our SimpleMatch. Green lines indicate correct matches, while red lines denote incorrect matches. The results are evaluated on PCK@0.05.}
    \label{fig:compare}
\end{figure*}

\textbf{Ablation on input resolution.} We further investigate optimal input resolution practices, as summarized in Table~\ref{tab:resolution}. The results show a notable performance improvement of 2.1\% when increasing the resolution from $256\times256$ to 320x320. The highest performance is achieved at an input resolution of 448x448, yielding a PCK@0.1 score of 74.7\%. Interestingly, within the range of 320x320 to 448x448, performance remains stable without significant fluctuations. This suggests that resolutions within this interval offer a practical trade-off between inference speed and accuracy, offering flexibility depending on computational constraints or application requirements.

\begin{table}[ht]
    \centering
    \begin{tabular}{l|c}
        \toprule
        Method & Training memory (MB)  \\
        \midrule
        Baseline & 21941 \\
        + window & 16287 \\
        + spare matching & 10675\\
        \bottomrule
    \end{tabular}
    \caption{Ablations on training efficiency.}
    \label{tab:memory}
\end{table}

\textbf{Training memory analysis.} As demonstrated in Table~\ref{tab:memory}, we validate the effectiveness of window-based localization and sparse matching in reducing training memory requirements. Our experimental results show that window-based localization alone reduces memory usage by 26\%. This approach not only constrains the computational context for target keypoint coordinate estimation but also achieves substantial memory savings. When further combined with sparse matching, the overall memory reduction reaches 51\%, marking a significant improvement in training efficiency. These findings suggest that our proposed methods can effectively alleviate memory constraints.

\subsection{Quantitative analysis}

To evaluate matching performance across different feature map resolutions, Figure~\ref{fig:match} provides an intuitive visualization of correspondence results at $16\times16$, $32\times32$, and $64\times64$ resolutions. The visualization reveals that, at the coarse $16\times16$ resolution, distinct source points will collapse into the same feature vector, causing distinct keypoints to match the same target keypoint. However, as the feature resolution increases to $32\times32$ and $64\times64$, a clear separation of target keypoints becomes observable, demonstrating that higher-resolution features better preserve spatial distinctions between neighboring descriptors.

We further compare our SimpleMatch with two SOTA semantic correspondence methods, ACTR and LPMFlow, focusing on fine-grained matching performance using the strict PCK@0.05 metric. As shown in Figure~\ref{fig:compare}, the results demonstrate that SimpleMatch SimpleMatch outperforms both ACTR and LPMFlow. These improvements can be attributed to its enhanced capability in handling adjacent keypoint distributions and learning more discriminative keypoint representations.

\section{Conclusion}


In this paper, we propose a simple yet effective framework, SimpleMatch, which addresses a fundamental limitation of keypoint feature fusion in semantic correspondence tasks. SimpleMatch consists of only a feature encoder and a lightweight decoder, complemented by a multi-scale loss function that enhances feature representation. Additionally, we incorporate sparse matching and window-based localization strategies to improve training efficiency, reducing  memory consumption by 51\%. Notably, SimpleMatch achieves SOTA performance while operating at a $3.3\times$ lower input resolution compared to existing methods, and it attains a throughput of 65 images per second, demonstrating remarkable computational efficiency. Due to its simplicity and effectiveness, we expect SimpleMatch to serve as a strong baseline for future work.

\section*{Acknowledgements}
This study was supported by the Jilin Scientific and Technological Development Program, China (Grant No. 20250102221JC) and the Interdisciplinary Innovation Center of Jilin University, China (Grant No. XKJCBY-2025018).

{
    \small
    \bibliographystyle{unsrt}
    \bibliography{references}
}

\end{document}